# CAVDETECT: A DBSCAN ALGORITHM BASED NOVEL CAVITY DETECTION MODEL ON PROTEIN'S STRUCTURE

SWATI ADHIKARI

*Department of Computer Science, The University of Burdwan, Golapbag, Purba Bardhaman, West Bengal, India, PIN: 713104*
swatidhkr@gmail.com

PARTHAJIT ROY*

*Department of Computer Science, The University of Burdwan, Golapbag, Purba Bardhaman, West Bengal, India, PIN: 713104*
roy.parthajit@gmail.com
*Corresponding Author

Abstract: Cavities on the structures of proteins are formed due to interaction between proteins and some small molecules, known as ligands. These are basically the locations where ligands bind with proteins. Actual detection of such locations is all-important to succeed in the entire drug design process. This study proposes a Voronoi Tessellation based novel cavity detection model that is used to detect cavities on the structure of proteins. As the atom space of protein's structure is dense and of large volumes and the DBSCAN (Density Based Spatial Clustering of Applications with Noise) algorithm can handle such type of data very well as well as it is not mandatory to have knowledge about the numbers of clusters (cavities) in data as priori in this algorithm, this study proposes to implement the proposed algorithm with the DBSCAN algorithm.

*Keywords*: CavDetect; Cavity Detection Techniques; Interaction between Protein & Ligand; Ligand; Ligand Binding Locations.

## 1 Introduction

The field which has gained much attention of the bio-researchers in recent years is drug design. The task of drug design is not a single one; it is composed of several steps, one of which is the recognition of ligand binding locations on protein's structures. Here, ligand means either protein or DNA or RNA or any other bio-molecule or metal ion or drug. Interaction between protein and ligand affects the structures and functions of proteins very much. For example, when drug interacts with protein at a specific location, the complex thus formed controls the spreading of drugs in the living body and changes the duration of action of drugs in the entire body. Protein-metal complexes also have much contribution in structural and functional changes of the proteins.

The locations where ligands bind on the structure of proteins are generally termed as cavities and these are some concave regions on the structure of proteins to which ligands bind. These cavities are responsible for performing many biological functions and are generally found either at the external surface or at the inside of the protein's structure. Cavities take various forms like pocket, cleft, channel, tunnel, void etc.[1] e.g., pockets & clefts are found at the surface of the proteins and voids are found inside the structures. Identification of cavities with these varieties of shapes is not a simple task.

The methods that have been used to detect cavities in earlier days are laboratory based. These traditional methods are very costly and take a long time to complete. To reduce the time & cost, computer-aided techniques are merged with the traditional methods. Simões et al. have reviewed different algorithms that are required for detection of cavities with various shapes along with some computational techniques.[1] Zhao et al. have also reviewed the same matter.[2] The authors have categorized the cavity detection algorithms as Geometry based, Evolutionary based and Energy based.[1] Presently, the cavity detection algorithms are grouped according to any combination of these three types of algorithms. These algorithms are further strengthened by the application of different tools of probability, statistics, machine learning, neural network as well as different tools of search and optimization. According to Zhao et al, the number of groups of algorithms for detecting cavities is four. These four groups of algorithms are based on the followings:

(i) three-dimensional structure of the proteins, (ii) similarities between sequence & structure templates, (iii) machine learning and (iv) deep learning.

Some of the methods that are based on (i) above consume more space & time than others. CavVis,[3] KVFinder,[4] Fpocket,[5] MSPocket,[6] FTSite,[7] SiteComp,[8] etc. are examples of methods that belong to this category.

The cavity detection methods that are based on (ii) above are also not fully advantageous. The main disadvantage associated with these methods are that the results generated with the template-based methods are highly influenced by the algorithm that is used to align sequences or structures. Sequence template similarity-based methods are not suitable for computing areas, volumes and shapes of detected ligand binding sites and structure template similarity-based methods also not suitable for predicting binding sites without having a three-dimensional structure. It is also becoming tough to predict any binding site regardless of any template with the same fold like the query protein that contains the ligands.[9]

The limitations of (i) and (ii) are overcome by integrating these methods with (iii) and (iv). Due to such effort, the efficiency and accuracy of the base methods also improve. Examples of (iii) and (iv) are MetaPocket,[10] ConCavity,[11] DeepSite,[12] DeepDrug3D,[13] DeepBind,[14] etc.

The advantages of the geometry and machine learning based cavity detection methods have been utilized in the proposed model of this study. The proposed algorithm starts with Voronoi Tessellation of the protein's atom space followed by clustering of Voronoi vertices to detect cavities by applying the density-based clustering algorithm DBSCAN. Next, few descriptors of these cavities are also computed. For evaluation of the proposed model, one custom dataset that contains 539 protein-magnesium complexes in 350 protein structures, which are taken from the database of Research Collaboratory for Structural Bioinformatics Protein Data Bank,[15] has been considered.

## 2 Materials & Methods

Present study discusses a novel cavity detection model that can find different types of cavities on the structures of proteins. This model combines the techniques of Voronoi Tessellation and clustering for the purpose, as it has been adopted by the Fpocket algorithm.[5] Here, the focus is to identify some empty areas on the structures of protein which are known to be as cavities.

For the fulfillment of this target, an alpha sphere-based concept has been adopted in the proposed model. An alpha sphere is formed by Voronoi Tessellation of an atom space. Through this decomposition several spheres are generated whose centres are the Voronoi vertices. These spheres contain at least four atoms in their boundary and interiors of these spheres are empty. When the protein's atom space is decomposed into number of Voronoi vertices, then the boundary atoms of alpha spheres represent the binding atoms of cavities. When these alpha spheres are grouped (clustered) based on the similarities between Voronoi vertices, probable groups of cavity binding atoms are identified.

Here, we can apply any clustering algorithm for grouping of alpha spheres (cavity binding atoms). But a wrong choice of clustering algorithm may generate clusters of alpha spheres that have very small sizes. These small clusters may be required to further merge with some other clustering algorithms to have larger clusters of cavity binding atoms as has happened in the Fpocket software[5] where three clustering steps have been applied for merging of the smaller clusters of alpha spheres. Although any number of clustering algorithms can be used to group the Voronoi vertices, due to this, time & space complexities will be increased with the application of more numbers of such algorithms. The output of such a merging process will also be connected to the success or drawback of each of the clustering algorithms.

Another point to be noted is that while detecting ligand binding sites, some algorithms like ConCavity[11] uses evolutionary information from the protein sequence. In the present study, the proposed model is constructed without having any evolutionary information.

Present paper proposes to apply the DBSCAN algorithm to group the alpha spheres. The advantage of using the DBSCAN algorithm over Partitional or Hierarchical clustering algorithms is that in the DBSCAN algorithm, it is not required to have the knowledge of numbers of groups (cavities) beforehand. Due to Voronoi Tessellation of protein's atom space, the numbers of alpha spheres generated will be huge in numbers (several thousands). The DBSCAN algorithm is capable of handling dense and large volumes of alpha spheres very well. This algorithm can find groups/clusters of different shapes from a large volume of dataset containing noise and outliers. In case of clusters having varying density, the DBSCAN algorithm does not work successfully. The DBSCAN algorithm also fails to find clusters of a dataset with high dimensional data.

**Proposed CavDetect Algorithm:**

The proposed algorithm for detection of cavities on the structure of protein is named as CavDetect algorithm. In this algorithm, at first Voronoi Tessellation is performed on all protein atoms. Due to this, the entire atom space is decomposed into numbers of regions which are very close to each other where each region represents a protein atom and three or more such regions intersect at the Voronoi vertex. Alpha sphere is a sphere whose centre is the Voronoi vertex and the number of such intersecting regions is more than three.

Next, to identify the cavity binding atoms, alpha spheres are grouped by applying the DBSCAN algorithm based on the nearness of the Voronoi vertices. Interiors of these groups of alpha spheres are empty. For clustering of alpha spheres, at first radii of these alpha spheres are measured and some alpha spheres are selected within a range of two threshold values (minimum & maximum cut). Next, all the Voronoi vertices (selected centres of the alpha spheres) are presented to the newly constructed model by following the DBSCAN algorithm. In this way, initial groups of alpha spheres are generated. The sizes of the groups which are less than a threshold value is merged to capture the larger cavities.

Now, it is found that not all cavities belong to the actual binding sites, the sites in which cavity atoms are bound with ligands present in protein structures. Some sites/cavities are also identified with the proposed CavDetect algorithm in which cavity atoms are bound with none of the ligands present in the structure concerned (non-binding site). Two criteria namely PocketPicker and Mutual Overlap criteria are applied to check whether the cavities that are detected with the CavDetect algorithm are actual binding sites or not, as these criteria are used to evaluate the algorithms PocketPicker[16] and Fpocket[5]. According to the PocketPicker criterion, in an actual binding site, the distance between the centroid of any binding site and any ligand atom lies within 4Å. As per Mutual Overlap criterion, in an actual binding site, 50% of the ligand atoms (at least) are found within 3Å of one alpha sphere (at least). Here, each ligand present in a structure is checked to see whether it has corresponding binding site or not.

After identification of actual binding sites and non-binding sites, some of the descriptors like (i) number of alpha spheres, (ii) density of the binding site, (iii) Polarity and charge score of the binding site, etc., which are defined in Fpocket,[5] are calculated for each of these two types of binding sites. Among these descriptors, (i) determines the size of any site, (ii) relates to the hydrophilicity character of the detected binding site and (iii) relates to the buried ness of the detected cavity where small and larger density values mean the site concerned is too buried and too exposed respectively.

**Flow Chart of the proposed CavDetect model:** Figure 1 depicts the flow chart of the proposed CavDetect model.

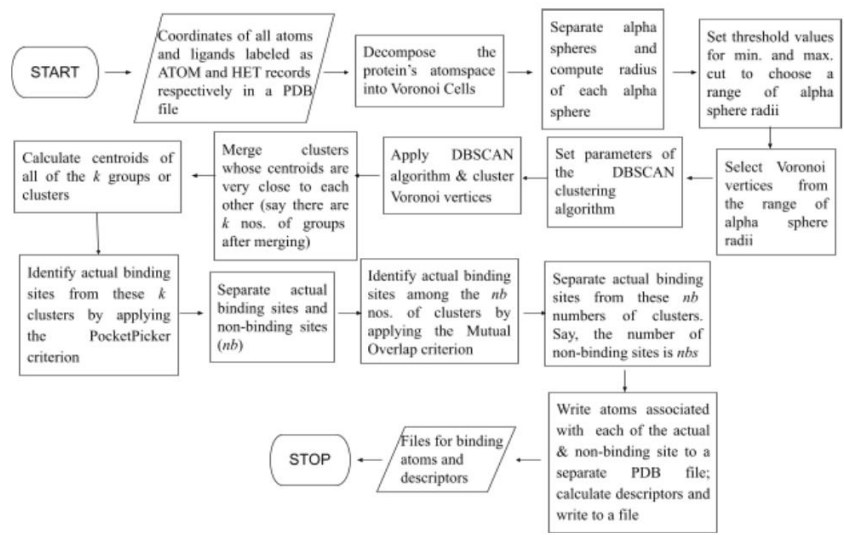

Figure 1: Flow Chart of the CavDetect Model

**Analysis of the CavDetect Model:**

Time complexities of computation of Voronoi Tessellation of '$x$' number of protein atoms and the application of DBSCAN algorithm on '$y$' number of Voronoi vertices are $O(x \log x)$ and $O(y^2)$ respectively.

Conversely in the Fpocket software,[5] Voronoi vertices are grouped by the application of the multiple clustering algorithms. In the first pass of the Fpocket algorithm, it is checked to see whether the Voronoi vertices are connected or not. If these vertices are connected through an edge, then any two vertices are merged if these are close enough. The single linkage algorithm is used in the second pass to further merging of these merged vertices. Here, $O(x^2)$ is the time complexity of the single linkage algorithm, '$x$' being the number of merged vertices. Next in the third pass of the Fpocket algorithm, to compare all vertices of one cluster to vertices of other clusters, the multiple linkage clustering algorithm is used. When a certain number of alpha spheres of two clusters are nearer, they are merged. So, along with complexities of $O(x \log x)$ and $O(x^2)$ which are required for Voronoi Tessellation and the single linkage algorithm, complexities of the first pass and the third pass are also taken into consideration in the Fpocket algorithm. That means, the complexity of the merging of Voronoi vertices in the CavDetect algorithm is lower than the Fpocket algorithm as well as lower than the algorithms that use multiple clustering algorithms.

**Dataset:** To test the CavDetect model, a custom dataset consisting of protein-magnesium complexes has been considered. To evaluate the proposed model, 350 protein structures of this custom dataset have been considered. Each ligand of this dataset has been considered. This dataset contains 3811 ligands (any kind), in total and 539 magnesium (Mg) ligands, in total. All these structures are taken from the database of Research Collaboratory for Structural Bioinformatics Protein Data Bank.[15]

## 3 Results & Discussion

The analysis of the results obtained after application of the CavDetect model on the test Mg-dataset is presented in this section. In this proposed model, the range of alpha sphere's radius is considered between 3Å and 5Å for inspection.

Following is some of the statistics related to the test Mg-dataset: no. of active ligands - 3334; no. of active sites - 2657; no. of non-active ligands - 477; no. of non-active sites - 15485; no. of active Mg sites - 494; no. of non-active Mg sites - 45

Above statistics shows that out of the 3811 ligands, which are present in the 350 test structures, the number of binding sites which are identified properly for 3334 ligands which are known as active ligands. The number of ligands for which no binding sites have been detected by the CavDetect algorithm is 477; these ligands are known as non-active ligands. Figure 2 shows that the structure 2vxt has 3 ligands of which two are CHLORIDE-ions and one is Mg-ion. Out of these 3 ligands, the binding sites for 2 CHLORIDE-ions have been detected successfully which are shown as the red & the blue-coloured pockets in Figure 2 whereas the binding site for the Mg-ion has not been detected. In this paper, PyMOL[17] has been used to create all figures.

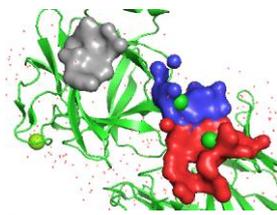

Figure 2: Structure 2vxt- Active & Non-Active Sites

Above statistics also show the presence of some non-active sites in the test structures which are 15485 in numbers. The gray-coloured pocket in Figure 2 shows one non-active site.

Sharing of binding sites is also seen through this model. Figure 3 depicts the sharing of 1 binding site by 2 ligands. In Figure 3, there are only 2 ligands GNP [PHOSPHOAMINOPHOSPHONIC ACID-GUANYLATE ESTER] and Mg-ion in the structure 2rgg which are sharing the same site.

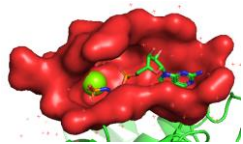

Figure 3: Structure 2rgg - Two ligands are sharing the same binding site

For the purpose of the evaluation of the CavDetect model, it has also been checked whether the CavDetect model can detect different types of cavities like clefts, channels, tunnels and voids. The structures 5it3 [Figure 4(a)], 6g2m [Figure 4(b)], 3k8y [Figure 4(c)] and 5cg5 [Figure 4(d)] have been considered for viewing cleft, channel, tunnel and void in respective binding sites.

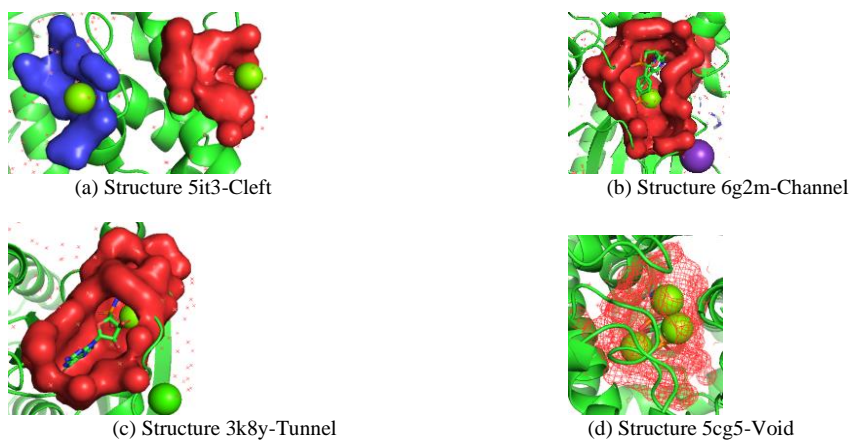

(a) Structure 5it3-Cleft  (b) Structure 6g2m-Channel

(c) Structure 3k8y-Tunnel  (d) Structure 5cg5-Void

Figure 4: View of Cleft, Channel, Tunnel & Void in Mg binding sites

It is also seen from the above statistics that out of the 539 Mg-sites, the proposed model has detected binding sites for 494 Mg-ions successfully. Some of the statistics relating to these 494 sites are presented in Table 1. In Table 1, all the 494 sites are divided into four categories - Cleft, Channel, Tunnel and Void. The average number of sites that come under each of these categories is also recorded.

For these sites, two descriptors viz., number of alpha spheres (ASs) and polarity score (PS) have been checked to compare size and hydrophilicity character of each site. The average numbers of sites that fall under these two types of descriptors are also recorded. To make the comparison process easier, the normalized values for the first and second descriptors are divided into six groups (smallest to largest) and five groups (least hydrophilic to most hydrophilic) respectively. Table 1 is presented below:

Table 1: Statistics regarding test Mg Binding sites, detected by the CavDetect Model

| Type of Pocket | Avg. No. of Mg Sites | Size (related to No. of AS)* [Avg. no. of sites in each group] | | | | | | Feature of Pockets (related to PS)# [Avg. no. of sites in each group] | | | | |
|---|---|---|---|---|---|---|---|---|---|---|---|---|
| | | I | II | III | IV | V | VI | A | B | C | D | E |
| **Cleft** | **0.85** | 0.08 | **0.90** | 0.02 | 0 | 0 | 0.003 | 0.04 | 0.20 | **0.53** | 0.09 | 0.15 |
| **Channel** | 0.06 | 0.04 | 0.32 | **0.57** | 0 | 0.04 | 0.04 | 0.11 | **0.39** | 0.29 | 0.18 | 0.04 |
| **Tunnel** | 0.03 | 0.06 | 0.31 | **0.44** | 0.13 | 0 | 0.06 | 0.13 | 0.25 | **0.50** | 0.06 | 0.06 |
| **Void** | 0.05 | 0.04 | **0.25** | **0.25** | **0.25** | 0.17 | 0.04 | 0.04 | **0.38** | 0.29 | 0.04 | 0.25 |

Categories under *: I - Smallest; II - Smaller; III - Small; IV - Large; V - Larger; VI - Largest

Categories under #: A - Least Hydrophilic; B - Less Hydrophilic; C - Hydrophilic; D - More Hydrophilic; E - Most Hydrophilic

From Table 1, following information has been obtained regarding Mg-binding sites:
(i) Mg-ion has a tendency to form clefts of varying sizes on the structures of proteins.
(ii) Sizes of maximum of these clefts are not so large.
(iii) Maximum of these clefts are hydrophilic in nature.
(iv) Although channels, tunnels and voids are also found, these are less in numbers.
(v) Overall sizes of maximum of these channels, tunnels and voids are small.
(vi) Hydrophilicity character of maximum channels and voids are not so high.
(vii) Hydrophilicity character of maximum tunnels is quite moderate.

## 4 Comparative Study

### 4.1 Comparison between the proposed CavDetect algorithm with the existing Fpocket[5] and ConCavity[11] algorithms

To evaluate the CavDetect model, the outputs generated with this algorithm are compared with the outputs generated with the other alpha shape-based software Fpocket.[5] Like Fpocket, the CavDetect algorithm can also detect different types of cavities. The proposed CavDetect model can also identify cavity atoms in .pdb format which can be visualized in PyMOL[17] and any other visualization software.

The output obtained with the CavDetect model for some structures is also compared with the output obtained with the algorithms of Fpocket[5] and ConCavity[11] softwares for the same structures. Following four structures of the test dataset are considered for comparison: 1q92, 3l8z, 4ql3 and 3fdo. Table 2 shows comparative study between the results obtained for these four structures. Table 2 is as the following:

**Table 2: Comparative study between the proposed algorithm and two existing algorithms**

| Struct. Name | No. of Ligands | Ligands Details [Residue ID, Chain Name, Residue No.] | No. of Active Ligands Detected by | | |
|---|---|---|---|---|---|
| | | | Fpocket | ConCavity | CavDetect |
| **1q92** | 3 | [DRM, A, 001], [GOL, A, 001], [MG, A, 003] | 1 | 2 | 3 |
| **3l8z** | 6 | [GNP, A, 201], [CA, A, 205], [MG, A, 202], [CA, A, 206, ], [CA, A, 203], [CA, A, 204] | 3 | 3 | 4 |
| **4ql3** | 2 | [GDP, A, 201], [MG, A, 202] | 0 | 2 | 2 |
| **3fdo** | 5 | [MG, B, 2], [MG, B, 3], [MG, A, 4], [MG, A, 5], [MG, A, 6] | 0 | 0 | 5 |

Statistics related to Table 2 are depicted in Figures 5 to 8.

Figure 5(c) shows that all the 3 ligands of the structure 1q92 have been successfully detected by the CavDetect model. But the Fpocket [Figure 5(a)] and the Concavity [Figure 5(b)] have detected the sites for only 1 and 2 ligands respectively.

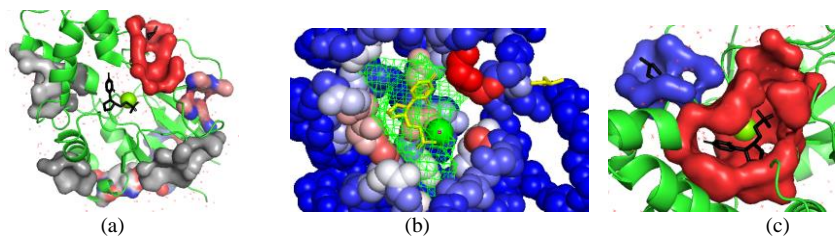

Figure 5: Structure 1q92: Outputs of algorithms - (a) Fpocket (c) ConCavity (d) CavDetect

From Figure 6(a), it is seen that the structure 3l8z has 6 ligands of which binding sites for 3 ligands have been successfully detected by the Fpocket [Figure 6(b)] and the Concavity [Figure 6(c)] whereas the CavDetect model has detected the same for only 4 ligands [Figures 6(d) to 6(f)].

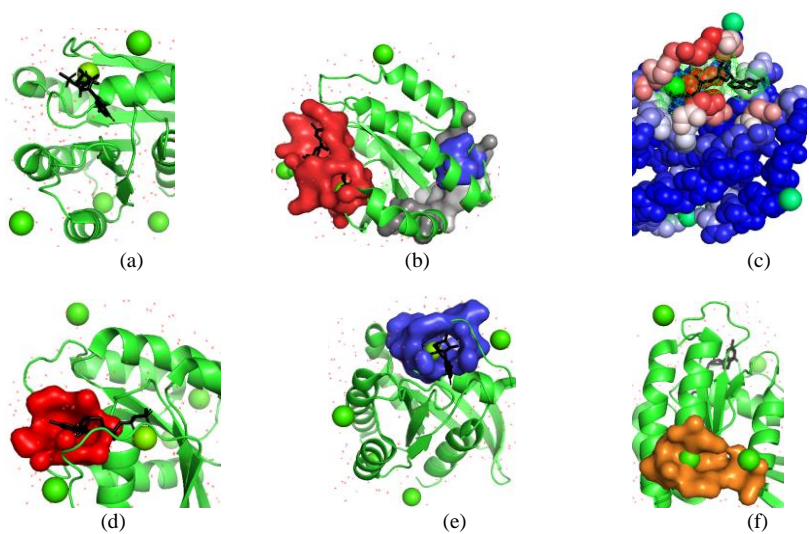

Figure 6: Structure 3l8z: (a) Only Ligands; Outputs of algorithms (b) Fpocket (c) ConCavity (d) CavDetect-Pocket1 (e) CavDetect-Pocket2 (f) CavDetect-Pocket3

Figure 7 shows that although the binding sites for all the 2 ligands of the structure 4ql3 remain undetected by the Fpocket, these ligands have been detected successfully by the ConCavity and the CavDetect model.

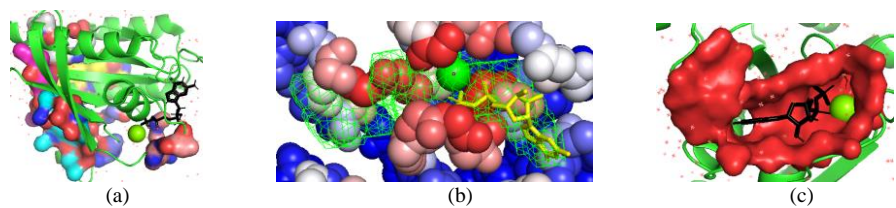

Figure 7: Structure 4ql3: Outputs of algorithms - (a) Fpocket (b) ConCavity (c) CavDetect

Figure 8 shows that although the binding sites for all the 5 ligands of the structure 3fdo remain undetected by Fpocket and ConCavity, these sites have been deleted successfully by the CavDetect model.

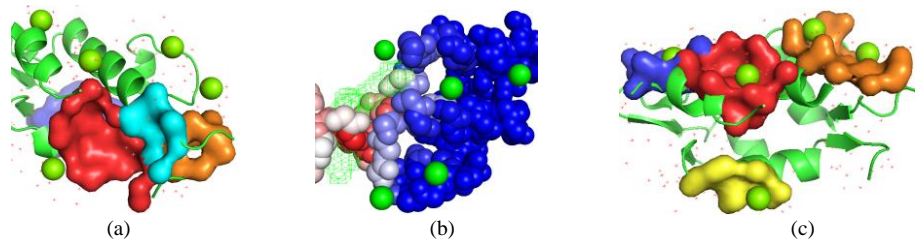

(a)　　　　　　　　(b)　　　　　　　　(c)

Figure 8: Structure 3fdo: Outputs of algorithms - (a) Fpocket (b) ConCavity (c) CavDetect

So, by comparing the Figures 5 to 8 and from Table 2 it is clear that the CavDetect model is capable of detecting binding sites as that of Fpocket and ConCavity wherein the CavDetect model sometimes outperforms the other two existing softwares.

Application of the DBSCAN clustering algorithm along with the outcome of Voronoi Tessellation in the proposed CavDetect model also removes the requirement of applying more than one clustering algorithms with the outcome of Voronoi Tessellation as well as removes the requirement of any evolutionary information in this algorithm.

This model also distinguishes between active and non-active sites. The provision for separate computation of descriptors for these two types of sites and generation of different information files in .txt format are incorporated in the CavDetect model. These information files also report about the active and the non-active ligands present in each structure.

## 4.2 Comparative Study of the CavDetect Model and the BioLiP Database[18]

As already stated, the proposed CavDetect model is developed based on a clustering algorithm. To compare the cavities detected through this model, 3 extrinsic clustering validity indices and 1 intrinsic validity index have been considered. For extrinsic validity indices, Precision, Recall, Accuracy indices are considered. For all these indices, ground truth labeled data is required. For the intrinsic validity index, Silhouette Score is considered. For this index, no ground truth labeled data is required.

To compare the CavDetect model, a total of 44 structures of the test Mg-dataset have been considered. The values for the above-mentioned 3 extrinsic validity indices are computed for these 44 structures. The output of the CavDetect model and the output obtained from the existing BioLiP[18] database have been compared. The structures of the test Mg-dataset contain different ions including metals and other materials like nucleic acids as ligands. For evaluation purposes, among these ligands, the binding sites for metals and nucleic acids are considered. For these 3 validity indices, as ground truth labeled data, the sites for metal-ions and the nucleic acids, detected by the MetBP[19] software, are considered. The Precision, Recall and Accuracy values are calculated (using these ground truth values) for the sites predicted with the CavDetect model and the BioLiP database. The sites that are detected with the MetBP, CavDetect and BioLiP are considered as positives and the sites that are not detected with these two algorithms & the database are considered as negatives. True positives, true negatives, false positives and false negatives values are computed based on these positive and negative values as well as the above mentioned three clustering validity indices are also computed for the CavDetect model and the BioLiP database. These values are shown in Table 3.

**Table 3:** Comparison of the CavDetect Model & the output obtained from the BioLiP[18] database

| Struc. Name | CavDetect | | | BioLiP | | | Struc. Name | CavDetect | | | BioLiP | | |
|---|---|---|---|---|---|---|---|---|---|---|---|---|---|
| | Prec | Rec | Acc (%) | Prec | Rec | Acc (%) | | Prec | Rec | Acc (%) | Prec | Rec | Acc (%) |
| 1r2q | 1 | 1 | 100 | 1 | 1 | 100 | 6gpo | 1 | 1 | 100 | **Div 0** | **0** | **0** |
| 5ph0 | 1 | 1 | 100 | 1 | 0.67 | 67 | 6gqs | 1 | 0.5 | 50 | **Div 0** | **0** | **0** |
| 6tan | 1 | 1 | 100 | 1 | 0.5 | 50 | 6hh2 | 1 | 1 | 100 | 1 | 0.5 | 50 |
| 6mkk | 1 | 1 | 100 | 1 | 1 | 100 | 6hmt | 1 | 1 | 100 | **Div 0** | **0** | **0** |
| 6san | 1 | 1 | 100 | 1 | 1 | 100 | 6hxu | 1 | 1 | 100 | 1 | 0.67 | 67 |
| 6ra7 | 1 | 1 | 100 | 1 | 1 | 100 | 6i3c | 1 | 1 | 100 | 1 | 1 | 100 |
| 6rhc | 1 | 0.67 | 67 | 0 | **Div 0** | 0 | 6pck | 1 | 1 | 100 | 1 | 1 | 100 |
| 6pjv | 0.33 | 1 | 33 | 0.33 | 1 | 33 | 6pcl | 0.33 | 1 | 33 | 0.33 | 1 | 33 |
| 6g6r | 1 | 0.75 | 75 | **Div 0** | **0** | **0** | 6rl3 | 1 | 1 | 100 | **Div 0** | **0** | **0** |
| 6ha3 | 0.67 | 1 | 67 | 1 | 0.5 | 67 | 6sa5 | 1 | 1 | 100 | 1 | 1 | 100 |
| 6had | 0.67 | 1 | 67 | 1 | 0.5 | 67 | 6t5b | 1 | 1 | 100 | 1 | 0.5 | 50 |
| 5vbm | 1 | 1 | 100 | 1 | 0.5 | 50 | 6g3r | 1 | 1 | 100 | 1 | 0.25 | 25 |
| 6p7z | 1 | 0.6 | 60 | 1 | 0.6 | 60 | 6iux | 1 | 1 | 100 | 1 | 1 | 100 |
| 6bcb | 1 | 1 | 100 | 1 | 1 | 100 | 6ms9 | 1 | 1 | 100 | 1 | 0.6 | 60 |
| 4l57 | 1 | 1 | 100 | 1 | 0.67 | 67 | 6n3d | 1 | 1 | 100 | No BioLiP Data | | |
| 5hob | 1 | 1 | 100 | No BioLiP Data | | | 6p0z | 1 | 1 | 100 | 1 | 0.5 | 50 |
| 6g2m | 1 | 0.5 | 50 | 1 | 0.5 | 50 | 6quw | 1 | 1 | 100 | 1 | 1 | 100 |
| 6g2n | 1 | 1 | 100 | 1 | 1 | 100 | 6y1j | 1 | 1 | 100 | 1 | 0.67 | 67 |
| 6g28 | 1 | 1 | 100 | 1 | 1 | 100 | 6bt1 | 1 | 1 | 100 | No BioLiP Data | | |
| 6g8k | 1 | 1 | 100 | **Div 0** | **0** | **0** | 6fbb | 1 | 1 | 100 | 1 | 0.5 | 50 |
| 6g8l | 1 | 0.5 | 50 | **Div 0** | **0** | **0** | 6fch | 1 | 1 | 100 | **Div 0** | **0** | **0** |
| 6gj5 | 1 | 1 | 100 | 1 | 1 | 100 | 6fcp | 1 | 1 | 100 | 1 | 0.5 | 50 |

Struc. - Structure; Prec - Precision; Rec - Recall; Acc - Accuracy; Div 0 - Division by 0

From Table 3 it is seen that the proposed CavDetect model can detect all nucleic acid and metal binding sites with full precision and recall values where accuracy is 100% (0.77, in average) but in case of the BioLiP database, the number of nucleic acid and metal binding sites which have been detected with full precision and recall values is less than the CavDetect model (0.3, in average). For the rest of the sites, in the CavDetect model, the accuracy is not below 33% and in the case of the BioLiP database, the accuracy is not below 25%.

In case of the BioLiP database, some of the entries for precision and recall values in Table 3 appear as Division by Zero. It appears when the BioLiP database fails to predict any binding site for the structure concerned. Figure 9 shows two such structures where the BioLiP database fails to predict any binding site for the metals present in these structures but the proposed CavDetect algorithm has successfully predicted the same. Figure 9 appear as follows:

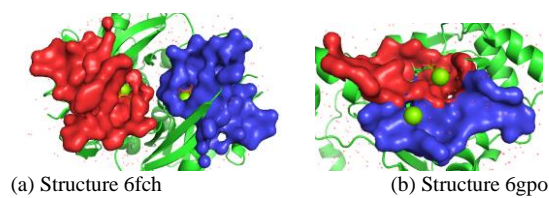

(a) Structure 6fch  (b) Structure 6gpo

Figure 9: Binding sites predicted with the CavDetect algorithm where the BioLiP database has failed to do so

So, from Table 3 and Figure 9, it is seen that in the sense of prediction of binding sites, the proposed CavDetect model is better than the existing BioLiP database for these 44 structures.

Now, it is required to check the exactness of these groups of alpha spheres as clustered by the CavDetect model. To check this, Silhouette Scores are computed for the clusters obtained with each of 44 structures by the CavDetect model. To compute the Silhouette Scores, along with metals and nucleic acids, other ligands present in each of the 44 structures have been taken into consideration. These Silhouette Scores for 44 structures are shown in Table 4 which is shown below:

**Table 4:** Silhouette Scores of the clusters generated with the CavDetect model

| Structure Name | CavDetect | Structure Name | CavDetect | Structure Name | CavDetect |
|---|---|---|---|---|---|
| 1r2q | 0.33 | 5hob | 0.35 | 6rl3 | 0.42 |
| 5ph0 | 0.32 | 6g2m | 0.37 | 6sa5 | 0.35 |
| 6tan | 0.38 | 6g2n | 0.36 | 6t5b | 0.36 |
| 6mkk | 0.40 | 6g28 | 0.41 | 6g3r | 0.35 |
| 6san | 0.40 | 6g8k | 0.39 | 6iux | 0.41 |
| 6ra7 | 0.43 | 6g8l | 0.44 | 6ms9 | 0.31 |
| 6rhc | 0.44 | 6gj5 | 0.42 | 6n3d | 0.44 |
| 6pjv | 0.37 | 6gpo | 0.39 | 6p0z | 0.35 |
| 6g6r | 0.44 | 6gqs | 0.40 | 6quw | 0.39 |
| 6ha3 | 0.33 | 6hh2 | 0.46 | 6y1j | 0.43 |
| 6had | 0.33 | 6hmt | 0.44 | 6bt1 | 0.41 |
| 5vbm | 0.38 | 6hxu | 0.39 | 6fbb | 0.44 |
| 6p7z | 0.35 | 6i3c | 0.39 | 6fch | 0.45 |
| 6bcb | 0.34 | 6pck | 0.35 | 6fcp | 0.45 |
| 4l57 | 0.39 | 6pcl | 0.36 | | |

Table 4 shows that the range of Silhouette Scores is between 0.3 and 0.5 which implies that all the groups have been correctly clustered with good quality.

## 5 Conclusion

The task of detection of ligand binding locations on the structure of proteins and to characterize those locations is not a simple one. It is a very important step in the context of drug designing. Present paper proposes a geometry-based algorithm that combines the clustering technique DBSCAN with it, called CavDetect. CavDetect can detect such locations with different shapes on protein's structures and computed some descriptors for the detected sites.

The CavDetect algorithm is evaluated by applying it on the above-mentioned Mg-dataset wherein all tables and figures show the satisfactory result. There is no requirement of application of various clustering algorithms to group the Voronoi vertices in the CavDetect algorithm. CavDetect also eliminates the need for evolutionary data and generates .pdb files for each of the detected ligand binding sites which helps in identification and visualization of the cavities; several text files are also generated consisting of computed descriptors of the detected ligand binding locations which may be useful for the researchers in the relevant field.

There is a scope for further improvement of the CavDetect algorithm by introducing the provision for ranking of the detected locations/sites. Instead of the DBSCAN algorithm, some different clustering algorithms can also be used to group the Voronoi vertices.

**Acknowledgement**

Authors express their gratitude to The University of Burdwan for the support and infrastructure they provided. For their online dataset, authors are also thankful to the Research Collaboratory for Structural Bioinformatics Protein Data Bank.